\documentclass[12pt]{l4dc2021} 

\usepackage{graphicx}
\usepackage{amsbsy}
\usepackage{array}
\usepackage{times}
\usepackage{mathtools}
\usepackage{graphicx}
\usepackage{blindtext}
\usepackage{bm}
\usepackage{todonotes}
\usepackage{multirow}
\usepackage{booktabs} 
\usepackage{amsbsy}
\usepackage{paralist} 
\usepackage{xspace} 
\usepackage{enumitem}
\usepackage{bbding} 
\usepackage{floatrow}

\usepackage{algorithm}
\usepackage{algorithmic}

\numberwithin{equation}{section}

\newcommand{\method}{\texttt{ST-SuEIR}\xspace}

\title[Bridging Physics-based and Data-driven modeling for  Learning Dynamical Systems]{Bridging Physics-based and Data-driven modeling for \\Learning Dynamical Systems }

\author{%
 \Name{Rui Wang} \Email{ruw020@ucsd.edu}\\
 \addr UC San Diego
 \AND
 \Name{Danielle Maddix} \Email{dmmaddix@amazon.com}\\
 \addr Amazon Research
  \AND
  \Name{Christos Faloutsos} \Email{faloutso@amazon.com}\\
  \addr Amazon and Carnegie Mellon University
  \AND
  \Name{Yuyang Wang} \Email{yuyawang@amazon.com}\\
  \addr Amazon Research
  \AND
 \Name{Rose Yu} \Email{roseyu@eng.ucsd.edu}\\
 \addr UC San Diego
}

\begin{document}
\maketitle
\vspace{-25pt}
\begin{abstract}%
How can we learn a dynamical system to make forecasts, when some variables are unobserved? For instance,
in COVID-19, we want to forecast the number of infected  and death cases but we do not know the count of susceptible and exposed people.  
While mechanics compartment models are widely-used in epidemic modeling, data-driven models are emerging for disease forecasting. We first formalize the learning of physics-based models as \texttt{AutoODE}, which leverages automatic differentiation to estimate the model parameters.
 Through a benchmark study on COVID-19 forecasting, we notice that physics-based mechanistic models significantly outperform deep learning.
%
%
Such performance differences highlight the generalization problem in dynamical system learning due to distribution shift. We identify two scenarios where distribution shift can occur:  changes in data domain  and changes in  parameter domain (system dynamics).  Through systematic experiments on several dynamical systems, we found that deep learning models fail to forecast well under  both scenarios.  While much research on distribution shift has focused on changes in data domain, our work calls attention to rethink generalization  for learning dynamical systems.
\end{abstract}

\vspace{-5pt}
\section{Introduction}
Dynamical systems \citep{strogatz2018nonlinear}  describe the evolution of phenomena occurring in nature, in which a differential equation, $d\bm{y}/dt = \bm{f}_{\bm{\theta}}(\bm{y}, t)$, models the dynamics of a $d$-dimensional state $\bm{y} \in \mathbb{R}^d$. Here $\bm{f}_{\bm{\theta}}$ is a non-linear operator parameterized by parameters $\bm{\theta}$.  Learning dynamical systems is to search a good model for a dynamical system in the hypothesis space guided by some criterion for performance. In this work, we study the  forecasting problem of predicting an sequence of future states $(\bm{y}_{k}, ..., \bm{y}_{k+q-1})$ given an sequence of historic states $(\bm{y}_{0}, ..., \bm{y}_{k-1})$. 

A plethora of work has been devoted to learning dynamical systems. When $\bm{f}_{\bm{\theta}}$ is known, physics-based methods based on numerical integration are commonly used for parameter estimation \citep{Houska2012}. \citet{mazier2, Ali2018Equ, Sirignano2018DGM} propose to directly solve $\bm{y}$ by approximating  $\bm{f}$ with neural networks that take the coordinates and time as input.
When $\bm{f}_{\bm{\theta}}$ is unknown and the data is abundant, data-driven methods are preferred. For example, deep learning (DL), especially deep sequence models \citep{Benidis2020NeuralFI, Sezer2020Survey, Flunkert2017DeepARPF, Rangapuram2018DeepSS} have demonstrated success in time series forecasting. In addition,  \cite{Wang2020TF, ayed2019learning, Wang2020symmetry, chen19} have developed hybrid DL models based on differential equations for spatiotemporal dynamics forecasting.

Our study was initially motivated by the need to forecast COVID-19 dynamics. While mechanics compartment models are widely used in epidemic modeling, data-driven deep learning models are emerging for disease forecasting. We formalize automatic differentiation for estimating mechanistic model parameters, which we name as \texttt{AutoODE}. Specifically, we use numerical integration  to generate state estimate  and minimize the difference between estimate and the ground truth with automatic differentiation \citep{Baydin2017AutomaticDI, Paszke17diff}. We also propose a novel compartmental model, \method, which obtains a 57.4\% reduction in mean absolute errors for 7-days ahead prediction.  We perform a comprehensive benchmark study of various methods to forecast the cumulative number of confirmed, removed and death cases.\footnote{Reproducibility: We open-source our code https://github.com/Rose-STL-Lab/AutoODE-DSL; the COVID-19 data we use is from John Hopkins Dataset https://github.com/CSSEGISandData/COVID-19.}
To our surprise, we found that physics-based models significantly outperform deep learning methods for COVID-19 forecasting.

To understand the inferior performance of DL, we experiment with several other dynamical systems: \textit{Sine}, \textit{SEIR}, \textit{Lotka-Volterra} and \textit{FitzHugh–Nagumo}. We observe that DL models suffer from distribution shift, often leading to poor generalization performance. In particular, two distribution shift scenarios may occur: changes in data distribution and changes in parameters that govern the dynamics of the system.  Our findings highlight the unique challenge of using DL models for learning dynamical systems. To bridge the gap between physics-based and data-driven modeling, we need robust methods that can deal with both distribution shift scenarios for the forecasting task.  To summarize, our contributions are the following:

\begin{compactenum}
    \item We study  dynamical systems for forecasting, even with unobserved variables. We formalize auto-differentiation for physics-based mechanistic models as \texttt{AutoODE} and propose a novel compartmental model for forecasting COVID-19.  Our method obtains a 57.4\% reduction in mean absolute error for 7-days ahead prediction compared to the best DL competitor.
    \item We perform a benchmark study of  physics-based mechanistic models and data-driven DL methods for predicting the COVID-19 dynamics among the cumulative confirmed, removed and death cases. We notice that physics-based models significantly outperform DL methods for COVID-19 forecasting, especially for the number of infected and removed cases.
    \item We expand our study  to other systems including \textit{Lotka-Volterra}, \textit{FitzHugh–Nagumo} and \textit{SEIR} dynamics. We observe that  the inferior performance of DL is mainly due to distribution shift. Many widely used DL models often  fail to learn the correct dynamics  when there is  distribution shift either in the data  or the  parameters of the dynamical system.
\end{compactenum}

\vspace{-5pt}
\section{Related Work}
\paragraph{Learning Dynamical System}
The seminal work by \cite{Brunton2015Sparse} proposed to solve ODEs by creating a dictionary of possible terms and applying sparse regression to select appropriate terms. But it assumes that the chosen library is sufficient. Physics-informed deep learning directly solves differential equations with neural nets given space $\bm{x}$ and time $t$ as input \citep{mazier2, Ali2018Equ, Sirignano2018DGM}. This type of methods cannot be used for forecasting since future $t$ would always lie outside of the training domain and neural nets cannot extrapolate to unseen domain \citep{Kouw2018domain, Amodei2016Safety}. Local methods, such as ARIMA and Gaussian SSMs \citep{Salina19gaussian, Rasmussen16Gaussian, Du16point} learn the parameters individually for each time series.  Hybrid DL models, e.g. \cite{ayed2019learning, Wang2020symmetry, chen19, Ayed2019LearningDS} integrate  differential equations in DL for temporal dynamics forecasting. 

\paragraph{Deep Sequence Models}
Since accurate numerical computation requires lots of manual engineering and theoretical properties may have not been well understood, deep sequence models have been widely used for learning dynamical systems. Sequence to sequence models and the Transformer, have an encoder-decoder structure that can directly map input sequences to output sequences with different lengths \citep{Vaswani2017AttentionIA, Wu2020Deep, Li2020Enhancing, Rangapuram2018DeepSS, Flunkert2017DeepARPF}. Fully connected neural networks can also be used autoregressively to produce multiple time-step forecasts \citep{Benidis2020NeuralFI, Lim2020TimeSF}. Neural ODE \citep{chen19} is based on the assumption that the data is governed by an ODE system and able to generate continuous predictions. When the data is spatially correlated, deep graph models, such as graph convolution networks and graph attention networks \citep{Petar2017Attention}, have also been used. 

\paragraph{Epidemic Forecasting}
Compartmental models are commonly used for modeling epidemics. 
\cite{Chen20SIR} proposes a time-dependent SIR model that uses ridge regression to predict the transmission and recovery rates over time. A potential limitation with this method is that it does not consider the incubation period and unreported cases.  
\cite{Pei20Initial} modified the compartments in the SEIR model into the subpopulation commuting among different places, and estimated the model parameters using iterated filtering methods. 
\cite{Wang20Survival} proposes a population-level survival-convolution method to model the number of infectious people as a convolution of newly infected cases and the proportion of individuals remaining infectious over time. 
\cite{zou20} proposes the \texttt{SuEIR} model that incorporates the unreported cases, and the effect of the exposed group on susceptibles. \cite{Davis20transmission} shows the importance of simultaneously modeling the transmission rate among the fifty U.S. states since transmission between states is significant.


\vspace{-5pt}
\section{Learning Dynamical Systems}
\subsection{Problem Formulation}
Denote $\bm{y} \in \mathbb{R}^d$ as  observed variables and $\bm{u} \in \mathbb{R}^p$ as the unobserved variables, we aim to learn a dynamical system given as
\begin{equation}\label{def}
\begin{dcases}
&\frac{d\bm{y}}{dt} = f_{\bm{\theta}}(t, \bm{y}, \bm{u})\\
&\frac{d\bm{u}}{dt} = g_{\bm{\theta}}(t, \bm{y}, \bm{u})\\
& \bm{y}(t_0) = \bm{y}_0 \\
& \bm{u}(t_0) = \bm{u}_0.
\end{dcases}
\end{equation}
In practice, we  have observations  $(\bm{y}_0, \bm{y}_1, ..., \bm{y}_{k-1})$ as inputs. The task of learning dynamical systems  is to learn $f_{\bm{\theta}}$ and $g_{\bm{\theta}}$,  and produce accurate forecasts $(\bm{y}_{k}, ..., \bm{y}_{k+q-1})$, where $q$ is called the forecasting horizon.

\subsection{Data-Driven Modeling} \label{sec_dl}
For data-driven models, we assume both $f$ and $g$ are unknown. We are given training and test samples  either as sliced sub-sequences from a long sequence (same parameters, different initial conditions) or independent samples from the  system  (different parameters, same initial conditions). In particular, let $p_\mathcal{S}$ be the training data distribution and $p_\mathcal{T}$ be the test data distribution. 
DL seeks a hypothesis $h \in \mathcal{H} : \mathbb{R}^{d \times k} \mapsto \mathbb{R}^{d \times q} $ that maps  a sequence of past values to  future values:
\begin{equation}\label{equ_task}
h(\bm{y}_{0}^{(i)}, ..., \bm{y}_{k-1}^{(i)}) = \hat{\bm{y}}_{k}^{(i)}, ..., \hat{\bm{y}}_{k+q-1}^{(i)}
\end{equation}
where  $(i)$ denotes individual sample, $k$ is the input length and $q$ is the output length. 


Following the standard statistical learning setting, a deep sequence model minimizes the  training loss $\hat{L_1}(h) = \frac{1}{n}\sum_{i=1}^n l(\bm{y}^{(i)}, h)$, where $\bm{y}^{(i)} = (\bm{y}_{0}^{(i)}, ..., \bm{y}_{k+q-1}^{(i)}) \sim p_\mathcal{S}$ is the $i^{th}$ training sample, $l$ is a loss function. For example, for square  loss, we have 
\[l(\bm{y}^{(i)}, h) = ||h(\bm{y}_{0}^{(i)}, ..., \bm{y}_{k-1}^{(i)}) - (\bm{y}_{k}^{(i)}, ..., \bm{y}_{k+q-1}^{(i)})||^2_2\]
The test error is given as $L_1(h) = \mathbb{E}_{\bm{y}\sim p_{\mathcal{T}}}[l(\bm{y}, h)]$. The goal is to achieve small test error $L_1(h)$ and small $|\hat{L_1}(h) - L_1(h)|$ indicates good generalization ability.

A fundamental difficulty of forecasting in dynamical system is the distributional shift that naturally occur in learning dynamical systems \citep{Kouw2018domain, Amodei2016Safety}. In forecasting, the data in the future $p_{\mathcal{T}}$ often lie outside the training domain $p_{\mathcal{S}}$, and requires methods to extrapolate to the unseen domain.  This is in contrast to classical machine learning theory, where generalization refers to model adapting to unseen data drawn from the same distribution \citep{Hastie17element, Poggio12theory}. 


\subsection{Physics-Based Modeling} \label{sec_autoode}
Physics-based modeling assumes we already have an appropriate system of ODEs to describe the underlying dynamics. We know the function $f$ and $g$, but not the parameters. We can use automatic differentiation to estimate the unknown parameters $\bm{\theta}$ and the initial values $\bm{u}_0$. We coin this procedure as \texttt{AutoODE}. Similar approaches  have been used in other papers \citep{Rackauckas2020UniversalDE, zou20} but have not been well formalized.  The main procedure is described in the algorithm 1. In the meanwhile, we need to ensure $\bm{u}$ and $\bm{y}$ have enough correlation that \texttt{AutoODE} can correctly learn all the parameters  based on the observable $\bm{y}$ only. If $\bm{u}$ and $\bm{y}$ are not correlated or loosely correlated, we may be not able to estimate $\bm{u}$ solely based on the observations of $\bm{y}$. 
 

\begin{table}[h!]
\centering
\begin{tabular}{p{13cm}}
\toprule
\textbf{Algorithm 1: \texttt{AutoODE}}\\
\midrule
0: Initialize the unknown parameters $\bm{\theta}, \bm{u}_0$ in Eqn. \ref{def} randomly. \\
1: Discretize Eqn. \ref{def} and apply 4-th order Runge Kutta (RK4) Method. \\
2: Generate estimation for $\bm{y}$: $(\hat{\bm{y}}_{0}, ..., \hat{\bm{y}}_{k})$ \\
3: Minimize the forecasting loss  with the Adam optimizer,\\
\quad $L_{2}(\bm{\theta}, \bm{u}_0) = \frac{1}{k}\sum_{i=0}^{k-1}||\hat{\bm{y}}_{i}(\bm{\theta}, \bm{u}, t) - \bm{y}_{i}(\bm{\theta}, \bm{u}, t)||^2.$\\
4: After convergence, use estimated $\hat{\bm{\theta}}, \hat{\bm{u}_0}$ and 4-th order Runge Kutta Method to \\
\quad generate final prediction, $(\bm{y}_{k}, ..., \bm{y}_{t+q-1})$. \\
\bottomrule
\end{tabular}
\label{algo:auto_ode}
\end{table}

{NeuralODE}  \citep{chen19} uses the adjoint method to differentiate through the numerical solver. Adjoint methods are more efficient in higher dimensional neural network models which require complex numerical integration. In our case, since we are dealing with low dimension ordinary differential equations and the RK4 is sufficient to generate accurate predictions. We can directly implement the RK4 in Pytorch and make it fully differentiable.


\vspace{-5pt}
\section{Case study: COVID-19 Forecasting}
We  benchmark different methods for predicting the COVID-19 dynamics among the cumulative confirmed, removed and death cases. We propose an extension of the  \texttt{SuEIR} compartmental model, and estimate the model parameters with \texttt{AutoODE}.
\begin{equation}\label{equ_sueird}
    \begin{dcases}
      &dS_i/dt = \textstyle -[\sum_j\beta_i(t) A_{ij}(I_j + E_j)S_i]/N_i,\\
      &dE_i/dt = \textstyle [\sum_j \beta_i(t) A_{ij}(I_j + E_j)S_i]/N_i - \sigma_i E_i,\\
      &dU_i/dt = (1-\mu_i) \sigma_i  E_i,\\
      &dI_i/dt = \mu_i\sigma_i E_i - \gamma_i I_i,\\
      &dR_i/dt = \gamma_i I_i,\\
      &dD_i/dt = r_i(t) dR_i/dt.\\
      &N_i = S_i + E_i + U_i + I_i + R_i
    \end{dcases}   
\end{equation}

\subsection{Spatiotemporal-SuEIR}
We present the \texttt{Spatiotemporal-SuEIR (ST-SuEIR)} model given in Eqn. \eqref{equ_sueird} and details are listed as below. We estimate the unknown parameters $\beta_i$, $\sigma_i$, $\mu_i$, and $\gamma_i$, which correspond to the transmission, incubation, discovery, and recovery rates, respectively. We also need to estimate unobserved variables $\bm{u} = \{S, E, U\}$ , the cumulative numbers of susceptibles, exposed and unreported cases, and predict the observed variables $\bm{y} = \{I, R, D\}$, cumulative numbers of infected, removed and death cases. The total population $N_i = S_i + E_i + U_i + I_i + R_i$ is assumed to be constant for each U.S. states $i$.


\paragraph{Low Rank Approximation to the Sparse Transmission Matrix $A_{ij}$} 
We introduce a sparse transmission matrix $A$ to model the transmission rate among the 50 U.S. states. $A$ is the element-wise product of the U.S. states adjacency matrix $M$ and the transmission matrix $C$, that is, $A = C \odot M \in \mathbb{R}^{50 \times 50}$. $M$ is a sparse 1-0 matrix that indicates whether two states are adjacent to each other. $C$ learns the transmission rates among the all 50 states from the data. $A = C \odot M$ means that we omit the transmission between the states that are not adjacent to each other. We make $C$ sparse with $M$ as $C$ contains too many parameters and we want to avoid overfitting. To further reduce the number of parameters and improve the computational efficiency to $\mathcal{O}(kn)$, we use a low rank approximation to generate the correlation matrix $C = B^TD$, where $B, D \in \mathbb{R}^{k\times n}$ for $k << n$.
	
\paragraph{Piecewise Linear Transmission Rate $\beta_i(t)$}
Most compartmental models assume the transmission rate $\beta_i$ is constant. For COVID-19, the transmission rate of COVID-19 changes over time due to government regulations, such as school closures and social distancing. Even though we focus on short-term forecasting (7 days ahead), it is possible that the transmission rate may change during the training period. Instead of a constant approximation to $\beta_i$, we use a piece-wise linear function over time $\beta_i(t)$, and set the breakpoints, slopes and biases as trainable parameters.

\paragraph{Death Rate Modeling: $r_i(t)$} The relationship between the numbers of accumulated removed and death cases can be close to linear, exponential or concave in different states. We assume the death rate $r_i(t)$  as a linear combination of $a_i t + b_i$ to cover both the convex and concave functions, where $a_i$ and $b_i$ are set as learnable parameters. 

\paragraph{Weighted Loss Function}
We set the unknown parameters in Eqn. \eqref{equ_sueird} as trainable, and apply \texttt{AutoODE} to minimize the following weighted loss function:
$$L(\bm{A}, \bm{\beta}, \bm{\sigma}, \bm{\mu},  \bm{\gamma}, \bm{r}) = \frac{1}{k}\sum_{t=0}^{k-1} w(t)\bigg[l(\hat{I}_t,I_t) + \alpha_1l(\hat{R}_t, R_t) + \alpha_2l(\hat{D}_t, D_t)\bigg],$$ 
with weights $\alpha_1, \alpha_2$ and loss function $l(\cdot, \cdot)$ which we will specify in the following section. We utilize these weights to balance the loss of the three states due to scaling differences, and also reweigh the loss at different time steps. We give larger weights to more recent data points by setting $w(t) = \sqrt{t}$. The constants, $\alpha_1, \alpha_2$ and $k$ are tuned on the validation set. 

\newcolumntype{P}[1]{>{\centering\arraybackslash}p{#1}}
\begin{table}[t!]
\small
\centering
\begin{center}
\begin{tabular}{P{2.5cm}P{0.8cm}P{0.8cm}P{0.8cm}P{0.8cm}P{0.8cm}P{0.8cm}P{0.8cm}P{0.8cm}P{0.8cm}P{0.8cm}}\toprule
\multirow{2}{*}{\textbf{MAE}} & \multicolumn{3}{c}{$07/13\sim 07/19$} &\multicolumn{3}{c}{$08/23\sim 08/29$} & \multicolumn{3}{c}{$09/06\sim 09/12$} \\
\cmidrule(rl){2-4} \cmidrule(rl){5-7} \cmidrule(rl){8-10}
& $I$ & $R$ & $D$ & $I$ & $R$ & $D$ & $I$ & $R$ & $D$ \\
\midrule
$\textbf{\texttt{FC}}$  & 8379 & 5330 & 257&  559 & 701 & \textbf{30} & 775 & 654 & \textbf{33} \\
\midrule
$\textbf{\texttt{Seq2Seq}}$ & 5172 & 2790 & \textbf{99} & 781 & 700 & 40  & 728 & 787 & 35  \\
\midrule
$\textbf{\texttt{Transformer}}$  & 8225 & 2937 & 2546 & 1282 & 1308 & 46 & 1301 & 1253 & 41 \\
\midrule
$\textbf{\texttt{NeuralODE}}$  & 7283& 5371& 173 & 682 & 661 & 43  & 858& 791 & 35 \\
\midrule
$\textbf{\texttt{GCN}}$  & 6843 & 3107 & 266 & 1066 & 923 & 55 & 1605 & 984 & 44 \\
\midrule
$\textbf{\texttt{GAT}}$  & 4155 & 2067 & 153 & 1003 & 898& 51 & 1065& 833& 40  \\
\midrule
$\textbf{\texttt{SuEIR}}$  & 1746 & 1984 & 136 &  639 & 778 & 39 & 888 & 637& 47   \\
\midrule
$\textbf{\texttt{ST-SuEIR}}$  & \textbf{818} & \textbf{1079} & 109 & \textbf{514} & \textbf{538} & 41 & \textbf{600} & \textbf{599} & 39\\
\bottomrule
\end{tabular}
\end{center}
\caption{\underline{Proposed \method} wins in predicting $I$ and $R$: 7-day ahead prediction MAEs on COVID-19 trajectories of accumulated number of infectious, removed and death cases. \vspace{-.3cm}}
\label{covid_mae}
\vspace{-5mm}
\end{table}

\begin{figure*}[t!]
	\centering
	\includegraphics[width= 0.32\textwidth]{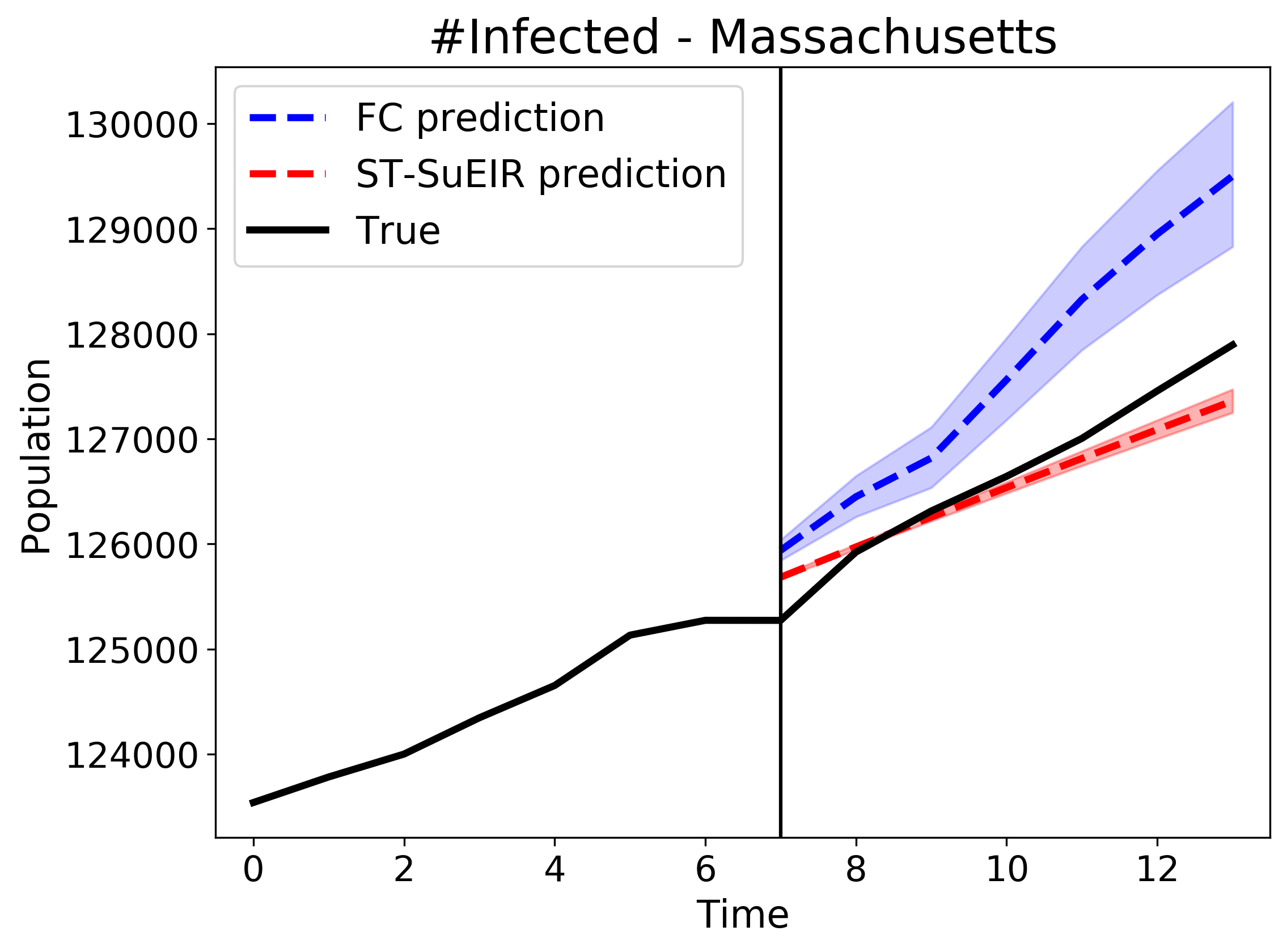}
	\includegraphics[width= 0.32\textwidth]{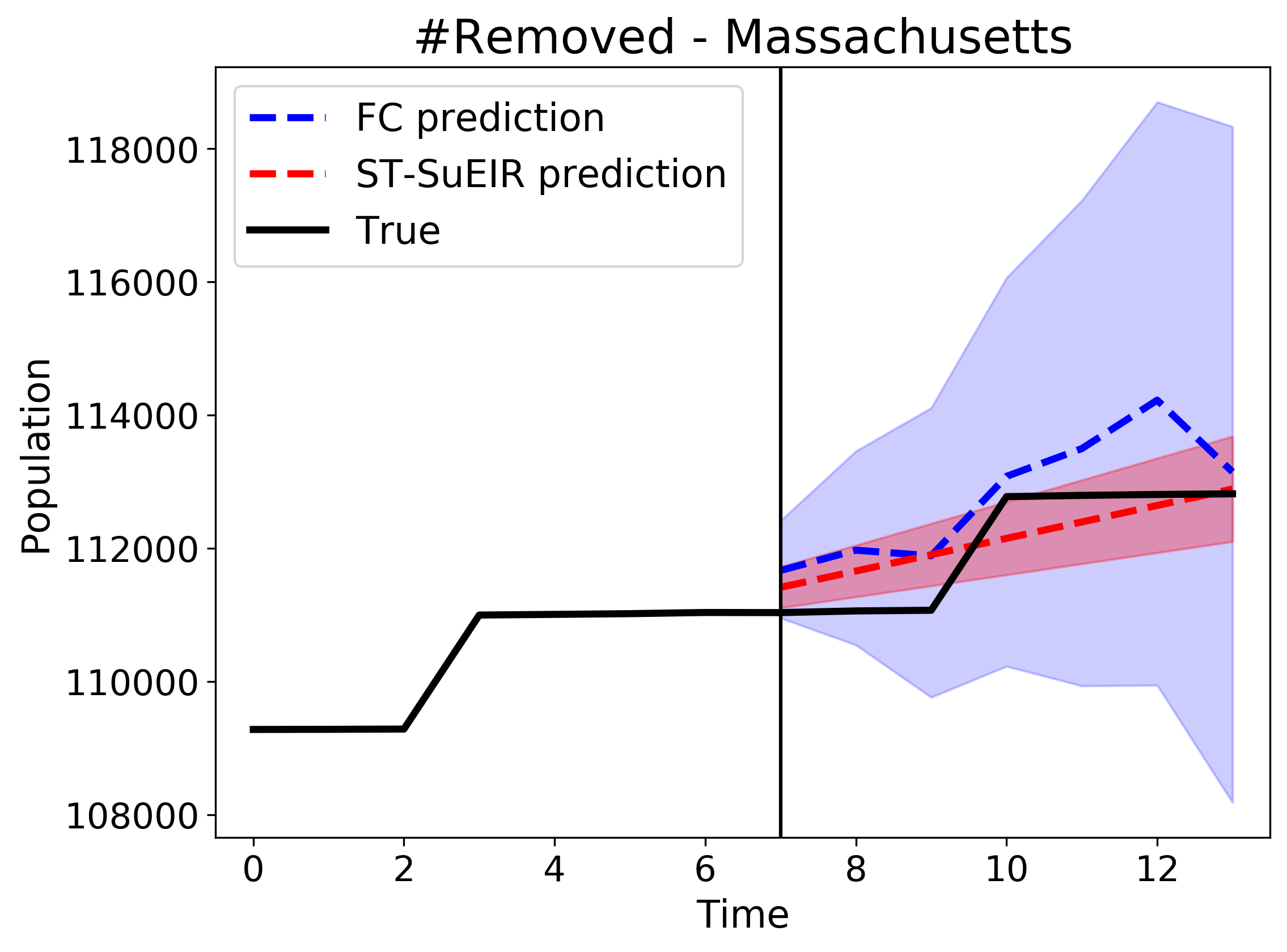}
	\includegraphics[width= 0.32\textwidth]{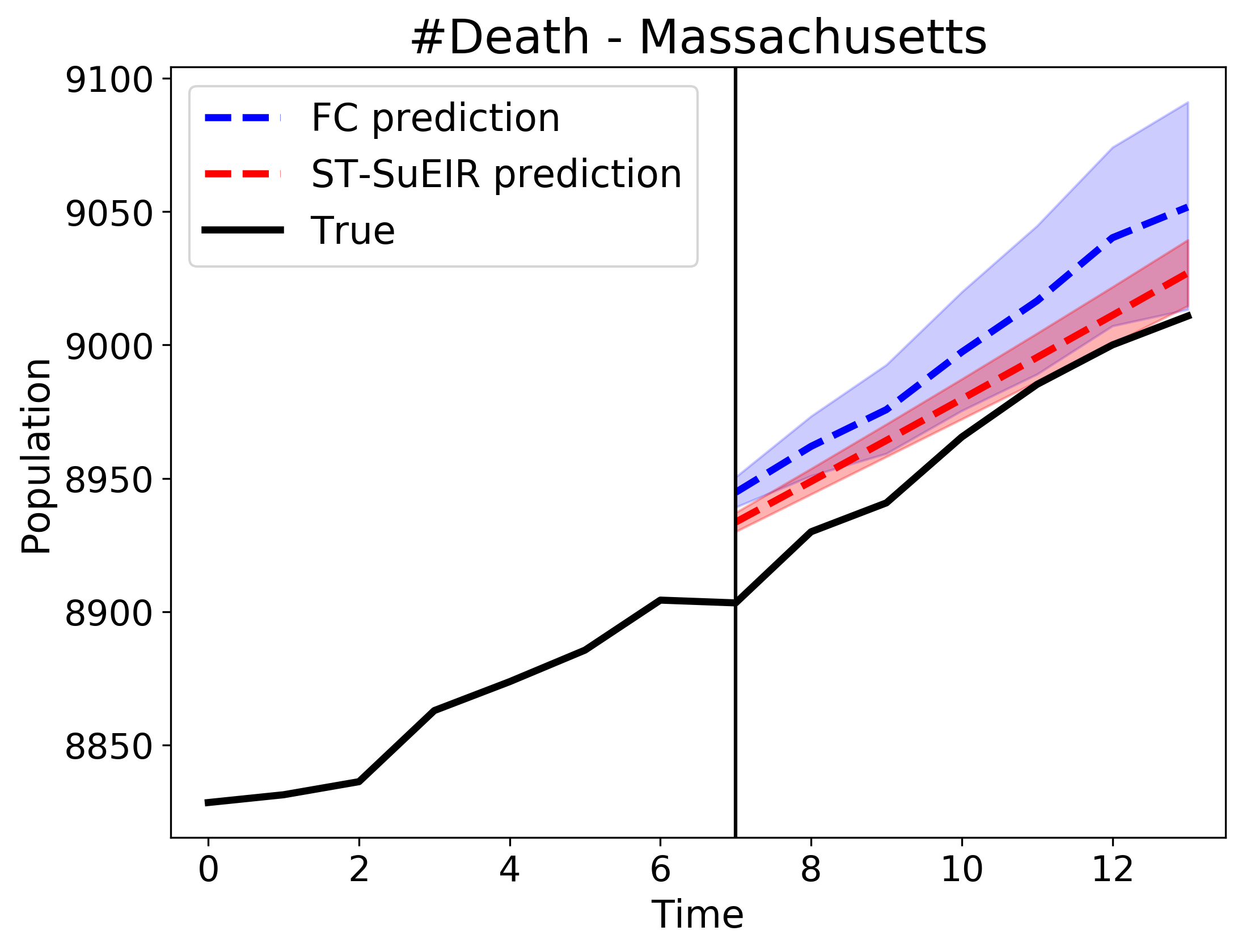}
	\caption{ \underline{Proposed \method wins}: $I$, $R$ and $D$ predictions for week $08/23\sim 08/29$ in Massachusetts by our proposed \method and the best performing DL model \texttt{FC}. \vspace{-0.25cm}}
	\label{fig:covid_pred_vis}
\end{figure*}

\subsection{Experiments on forecasting COVID-19 Dynamics} 
We use the COVID-19 data from Apr 14 to Sept 12 provided by Johns Hopkins University \citep{covid19jhu}. It contains the cumulative numbers of infected ($I$), recovered ($R$) and death ($D$) cases. 

We investigate  six DL models on forecasting COVID-19 trajectories: sequence to sequence with LSTMs (\texttt{Seq2Seq}), \texttt{Transformer}, autoregressive fully connected neural nets (\texttt{FC}), \texttt{NeuralODE}, graph convolution networks (\texttt{GCN}) and graph attention networks (\texttt{GAT}). We standardize $I$, $R$ and $D$ time series of each state individually to avoid one set of features dominating another. We use sliding windows to generate samples of sequences before the target week and split them into training and validation sets. We perform exhaustive search to tune the hyperparameters on the validation set. 

We also compare \texttt{SuEIR} \citep{zou20} and \texttt{ST-SuEIR}. We rescale the trajectories of the number of cumulative cases of each state by the population of that state. We use the quantile regression loss \citep{MQCNN2018Wen} for all models. All the DL models are trained to predict the number of daily new cases instead of the number of cumulative cases because we want to detread the time series, and put the training and test samples in the same approximate range. All experiments were conducted on Amazon Sagemaker \citep{sagemaker}. 

Table \ref{covid_mae} shows the 7-day ahead forecasting mean absolute errors of three features $I$, $R$ and $D$ for the weeks of July 13, Aug 23 and Sept 6. We can see that \method overall performs better than \texttt{SuEIR} and all the DL models. \texttt{FC} and \texttt{Seq2Seq} have better prediction accuracy of death counts but all DL models have much bigger errors on the prediction of week July 13. Figure \ref{fig:covid_pred_vis} visualizes the 7-day ahead COVID-19 predictions of $I$, $R$ and $D$ in Massachusetts by \method and the best performing DL model, \texttt{FC}. The prediction by \method is closer to the target and has smaller confidence intervals. This demonstrates the effectiveness of our hybrid model, as well as the benefits of our novel compartmental model design.  


\vspace{-5pt}
\section{Generalization in Learning Dynamical Systems}
We explore the potential reasons behind the inadequate performance of DL models on COVID-19 forecasting. It is known that distribution shift often leads to poor generalization in DL. A model's performance deteriorates quickly when the test data distribution is different from training.     We  explore two distribution shift scenarios:  changes in the data  where the observation domain differs; and changes in parameters  where the dynamics of  the  system in training and  test differs. 

\subsection{Dynamical Systems}\label{syn_dynamics}

Apart from COVID-19 trajectories, we also investigate the following three non-linear dynamics. 
\paragraph{Lotka-Volterra (\textit{LV})} system \citep{Ahmad1993OnTN} of Eqn.\eqref{equ-lv} describes the dynamics of biological systems in which predators and preys interact, where $d$ denotes the number of species interacting and $p_i$ denotes the population size of species $i$ at time $t$.  The unknown parameters $r_i \ge 0$, $k_i \ge 0$ and $A_{ij}$ denote the intrinsic growth rate of species $i$, the carrying capacity of species $i$ when the other species are absent, and the interspecies competition between two different species, respectively. 

\paragraph{FitzHugh–Nagumo (\textit{FHN})} \cite{FitzHugh} and, independently, \cite{Nagumo} derived the Eqn.\eqref{equ-fhn} to qualitatively describe the behaviour of spike potentials in the giant axon of squid neurons. The system describes the reciprocal dependencies of the voltage $x$ across an axon membrane and a recovery variable $y$ summarizing outward currents. The unknown parameters $a$, $b$, and $c$ are dimensionless and positive, and $c$ determines how fast $y$ changes relative to $x$.

\paragraph{SEIR} system of Eqn.\eqref{equ-seir} models the spread of infectious diseases \citep{Tillett1992Dynamics}. It has four compartments: Susceptible ($S$) denotes those who potentially have the disease, Exposed ($E$) models the incubation period, Infected ($I$) denotes the infectious who currently have the disease, and Removed/Recovered ($R$) denotes those who have recovered from the disease or have died.  The total population $N$ is assumed to be constant and the sum of these four states.  The unknown parameters $\beta$, $\sigma$ and $\gamma$ denote the transmission, incubation, and recovery rates, respectively.

\begin{minipage}[t]{0.29\textwidth}
\begin{align}\label{equ-lv}
\frac{dp_i}{dt} = & r_ip_i\Big(1 - \frac{\sum_{j=1}^d A_{ij}p_j}{k_i}\Big), \nonumber\\
i =& 1, 2, \dots, d.
\end{align}
\end{minipage}
\begin{minipage}[t]{0.3\textwidth}
\begin{equation}\label{equ-fhn}
\begin{dcases}
&\frac{dx}{dt} = c(x + y - \frac{x^3}{3}),\\
&\frac{dy}{dt} = -\frac{1}{c}(x + by - a).
\end{dcases}
\end{equation}
\end{minipage}
\begin{minipage}[t]{0.36\textwidth}
\begin{equation}\label{equ-seir}
    \begin{dcases}
      &dS/dt = -\beta SI/N,\\
      &dE/dt = \beta SI/N - \sigma E,\\
      &dI/dt = \sigma E - \gamma I,\\
      &dR/dt = \gamma I,\\
      & N = S + E + I + R.
    \end{dcases}
\end{equation}
\vspace{-5mm}
\end{minipage}

\subsection{Distribution Shift: Interpolation vs. Extrapolation}
We define $p_\mathcal{S}$ and $p_\mathcal{T}$ as the training and the test data distributions. And the $\theta_\mathcal{S}$ and $\theta_\mathcal{T}$ denote \textit{parameter} distributions of training and test sets, where the \textit{parameter} here refers to the coefficients and the initial values of dynamical systems. A distribution is a function that map a sample space to the interval [0,1] if it a continuous distribution, or a subset of that interval if it is a discrete distribution. The domain of a distribution $p$, i.e. $\text{Dom}(p)$, refers to the set of values (sample space) for which that distribution is defined.

We define two types of interpolation and extrapolation tasks. Regarding the data domain, we define a task as an interpolation task when the data domain of the test data is a subset of the domain of the training data, i.e., $\text{Dom}(p_{\mathcal{T}}) \subseteq \text{Dom}(p_{\mathcal{S}})$, and then extrapolation occurs $\text{Dom}(p_{\mathcal{T}}) \not\subseteq \text{Dom}(p_{\mathcal{S}})$. Regarding the \textit{parameter} domain, an interpolation task indicates that $\text{Dom}(\theta_{\mathcal{T}}) \subseteq \text{Dom}(\theta_{\mathcal{S}})$, and an extrapolation task indicates that $\text{Dom}(\theta_{\mathcal{T}}) \not\subseteq \text{Dom}(\theta_{\mathcal{S}})$. Many machine learning setups focus on the interpolation tasks. The extrapolation tasks correspond to the situations with distribution shift.

\subsection{Scenario 1: unseen data in the different data domain} \label{data_domain}
Through a simple experiment on learning the \textit{Sine} curves, we show deep sequence models have poor generalization on extrapolation tasks regarding the data domain, i.e. $\text{Dom}(p_{\mathcal{T}}) \not\subseteq \text{Dom}(p_{\mathcal{S}})$. Specifically, we generate 2k \textit{Sine} samples of length 60 with different frequencies and phases, and randomly split them into training, validation and interpolation-test sets. The extrapolation-test set is the interpolation-test set shifted up by 1. We investigate four models, including \texttt{Seq2Seq} (sequence to sequence with LSTMs), \texttt{Transformer}, \texttt{FC} (autoregressive fully connected neural nets) and \texttt{NeuralODE}. All models are trained to make 30 steps ahead prediction given the previous 30 steps. 

\begin{figure}
\begin{floatrow}
\ffigbox[0.65\textwidth]
{%
\includegraphics[width=0.48\linewidth]{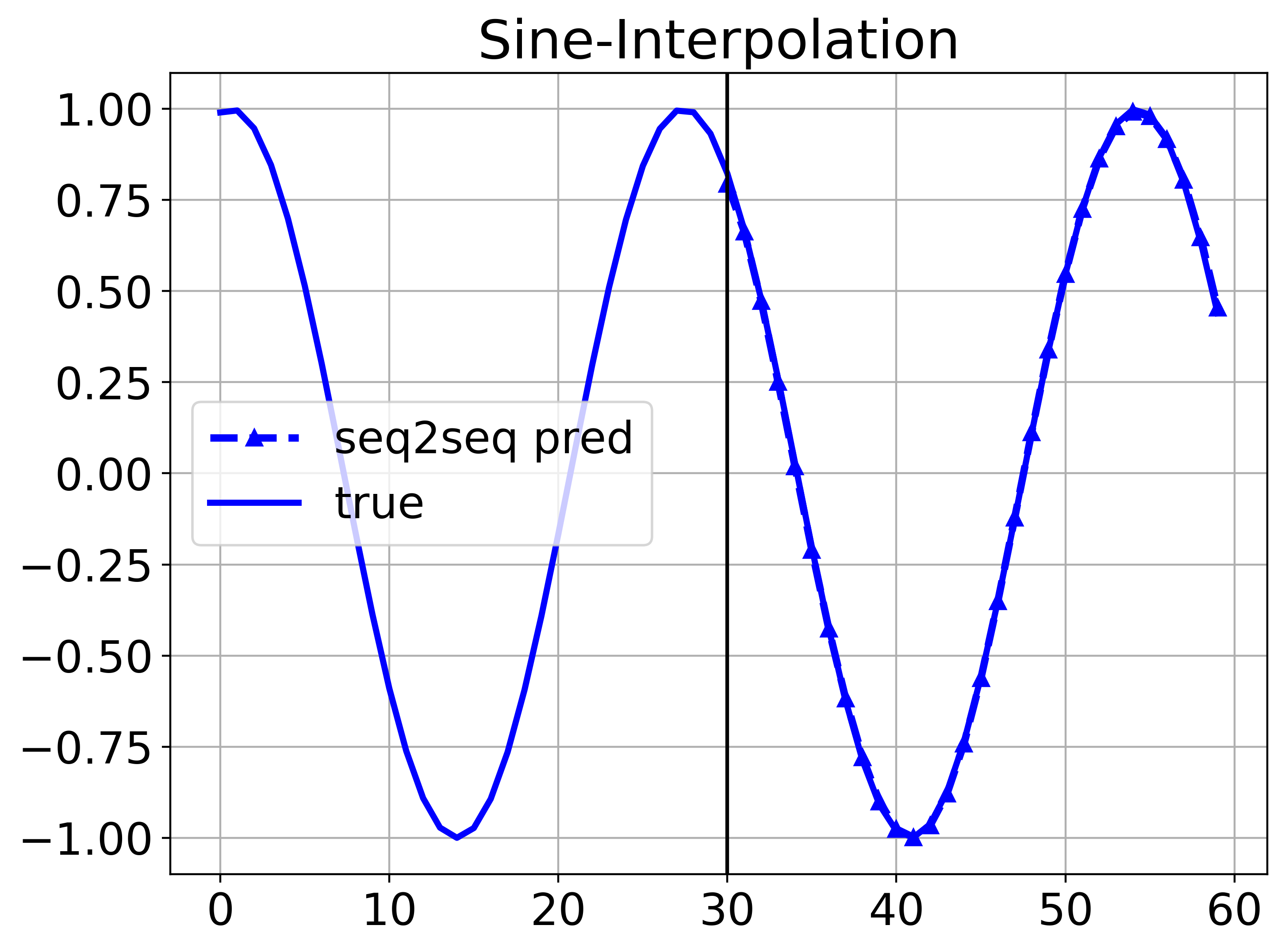}
\includegraphics[width=0.48\linewidth]{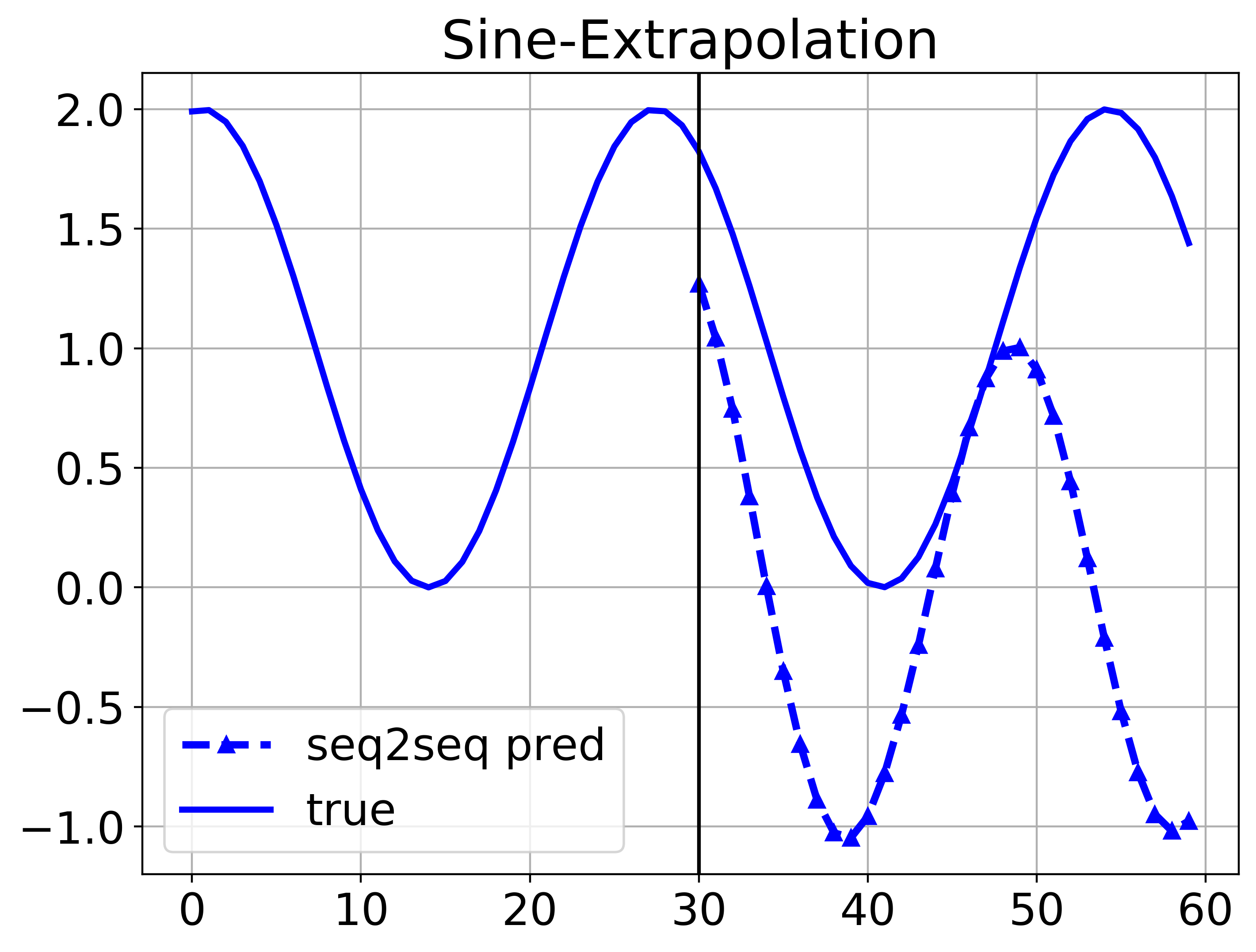}
}{
\caption{\texttt{Seq2Seq} predictions on an interpolation (left) and an extrapolation (right) test samples of Sine dynamics, the vertical black line in the plots separates the input and forecasting period.}\label{sine_fig_inter_extra}
}
\capbtabbox[0.34\textwidth]
{
 \caption{RMSEs of the interpolation and extrapolation tasks of Sine dynamics.}\label{sine_inter_extra}
\vspace{-.25cm}}
{
\footnotesize
 \begin{tabular}{P{1.9cm}|P{0.8cm}P{0.8cm}}\toprule
 \textbf{RMSE} & \texttt{Inter} & \texttt{Extra} \\
 \midrule
 \textbf{\texttt{Seq2Seq}} & 0.012 & 1.242   \\
 \midrule
 \textbf{\texttt{Auto-FC}}  & 0.009 & 1.554  \\
 \midrule
 \textbf{\texttt{Transformer}}  & 0.016 & 1.088  \\
 \midrule
 \textbf{\texttt{NeuralODE}} & 0.012 & 1.214 \\ 
 \bottomrule
 \end{tabular}
}
\end{floatrow}
\vspace{-3mm}
\end{figure}

Table \ref{sine_inter_extra} shows that all models have substantially larger errors on the extrapolation test set. Figure \ref{sine_fig_inter_extra} shows \texttt{Seq2Seq} predictions on an interpolation (left) and an extrapolation (right) test samples. We can see that \texttt{Seq2Seq} makes accurate predictions on the interpolation-test sample, while it fails to generalize when the same samples are shifted up only by 1.

\newfloatcommand{capbtabbox}{table}[][\FBwidth]

\newcolumntype{P}[1]{>{\centering\arraybackslash}p{#1}}
\subsection{Scenario 2: unseen data with different system parameters} \label{param_domain}
Even when $\text{Dom}(p_{\mathcal{T}}) \subseteq \text{Dom}(p_{\mathcal{S}})$, deep sequence models can still fail to learn the correct dynamics if there is a distributional shift in the parameter domain, i.e., $\text{Dom}(\theta_{\mathcal{T}}) \not\subseteq \text{Dom}(\theta_{\mathcal{S}})$. For example, the recovery rate $\gamma$ in the epidemic may increase as more people are vaccinated. A tighter social distancing policy would lead to smaller contact rate $\beta$. Thus, there is a great chance that new test samples are outside of the parameter domain of training data. In that case, the DL models would not make accurate prediction for COVID-19. 

For each of the three dynamics in section \ref{syn_dynamics}, we generate 6k synthetic time series samples with different system parameters and initial values. The training/validation/interpolation-test sets for each dataset have the same range of system parameters while the extrapolation-test set contains samples from a different range. Table \ref{test_param} shows the \textit{parameter} distribution of test sets. For each dynamics, we perform two experiments to evaluate the models' extrapolation generalization ability on initial values and system parameters. All samples are normalized so that $\text{Dom}(p_{\mathcal{T}}) = \text{Dom}(p_{\mathcal{S}})$. 

\begin{table*}[htb!]
\footnotesize
\caption{The initial values and system parameters ranges of interpolation and extrapolation test sets.}\label{test_param}
\begin{tabular}{P{1.2cm}|P{2.5cm}P{2.5cm}|P{2.5cm}P{2.5cm}}\toprule
\multirow{2}{*}{} & \multicolumn{2}{c}{System Parameters} &\multicolumn{2}{c}{Initial Values}\\
\cmidrule(rl){2-3} \cmidrule(rl){4-5} 
& Interpolation & Extrapolation & Interpolation & Extrapolation  \\
\midrule
$\textit{LV}$  & $\bm{k} \sim U(0, 250)^4$ & $\bm{k} \sim U(250, 300)^4$ & $\bm{p_0} \sim U(30, 200)^4$ & $\bm{p_0} \sim U(0, 30)^4$ \\
\midrule
$\textit{FHN}$  & $c \sim U(1.5, 5)$ & $c \sim U(0.5, 1.5)$ & $x_0 \sim U(2, 10)$ & $x_0 \sim U(0, 2)$  \\
\midrule
$\textit{SEIR}$  & $\beta \sim U(0.45, 0.9)$ & $\beta \sim U(0.3, 0.45)$ & $I_0 \sim U(30, 100)$ & $I_0 \sim U(10, 30)$\\
\bottomrule
\end{tabular}
\vspace{-3mm}
\end{table*}

\begin{table}
\footnotesize
\begin{tabular}{P{2cm}|P{0.6cm}P{0.6cm}P{0.6cm}P{0.6cm}|P{0.6cm}P{0.6cm}P{0.6cm}P{0.6cm}|P{0.5cm}P{0.5cm}P{0.5cm}P{0.5cm}}\toprule
\multirow{3}{*}{\textbf{RMSE}}& \multicolumn{4}{c|}{$\textit{LV}$} & \multicolumn{4}{c|}{$\textit{FHN}$} & \multicolumn{4}{c}{$\textit{SEIR}$}\\
\cmidrule(rl){2-5} \cmidrule(rl){6-9} \cmidrule(rl){10-13}
& \multicolumn{2}{c}{$\bm{k}$} & \multicolumn{2}{c|}{$\bm{p_0}$} & \multicolumn{2}{c}{$c$} & \multicolumn{2}{c|}{$x_0$} & \multicolumn{2}{c}{$\beta$} & \multicolumn{2}{c}{$I_0$} \\\cmidrule(rl){2-3} \cmidrule(rl){4-5} \cmidrule(rl){6-7} \cmidrule(rl){8-9} \cmidrule(rl){10-11} \cmidrule(rl){12-13}
& \texttt{Int} & \texttt{Ext} & \texttt{Int} & \texttt{Ext} & \texttt{Int} & \texttt{Ext} & \texttt{Int} & \texttt{Ext} & \texttt{Int} & \texttt{Ext} & \texttt{Int} & \texttt{Ext} \\
\midrule
\textbf{\texttt{Seq2Seq}} & \textbf{0.050} & 0.215 & 0.028 & 0.119 & 0.093 & 0.738 & 0.079 & 0.152 & 1.12 & 4.14 & 2.58 & 7.89\\
\midrule
\textbf{\texttt{FC}}  & 0.078 & 0.227 & 0.044 & 0.131 & \textbf{0.057} & 0.402 & \textbf{0.057} & 0.120 & 1.04 & 3.20 & 1.82 & 5.85\\
\midrule
\textbf{\texttt{Transformer}}  & 0.074 & 0.231 & 0.067 &  0.142 & 0.102 & 0.548 & 0.111 & 0.208 & 1.09 & 4.23 & 2.01 & 6.13\\
\midrule
\textbf{\texttt{NeuralODE}}  & 0.091 & 0.196 & 0.050 & 0.127  & 0.163 & 0.689 & 0.124 & 0.371& 1.25 & 3.27 & 2.01  &  5.82\\
\midrule
\textbf{\texttt{AutoODE}} & 0.057 & \textbf{0.054} & \textbf{0.018} & \textbf{0.028}  & 0.059 & \textbf{0.058} & 0.066 & \textbf{0.069} & \textbf{0.89} & \textbf{0.91} & \textbf{0.96}  & \textbf{1.02}\\
\bottomrule
\end{tabular}
\caption{RMSEs on initial values and system parameter interpolation and extrapolation test sets.}\label{test_rmse}
\vspace{-5mm}
\end{table}

\begin{figure}[t!]
\begin{floatrow}
\ffigbox[0.49\textwidth]
{%
\includegraphics[width=0.24\textwidth]{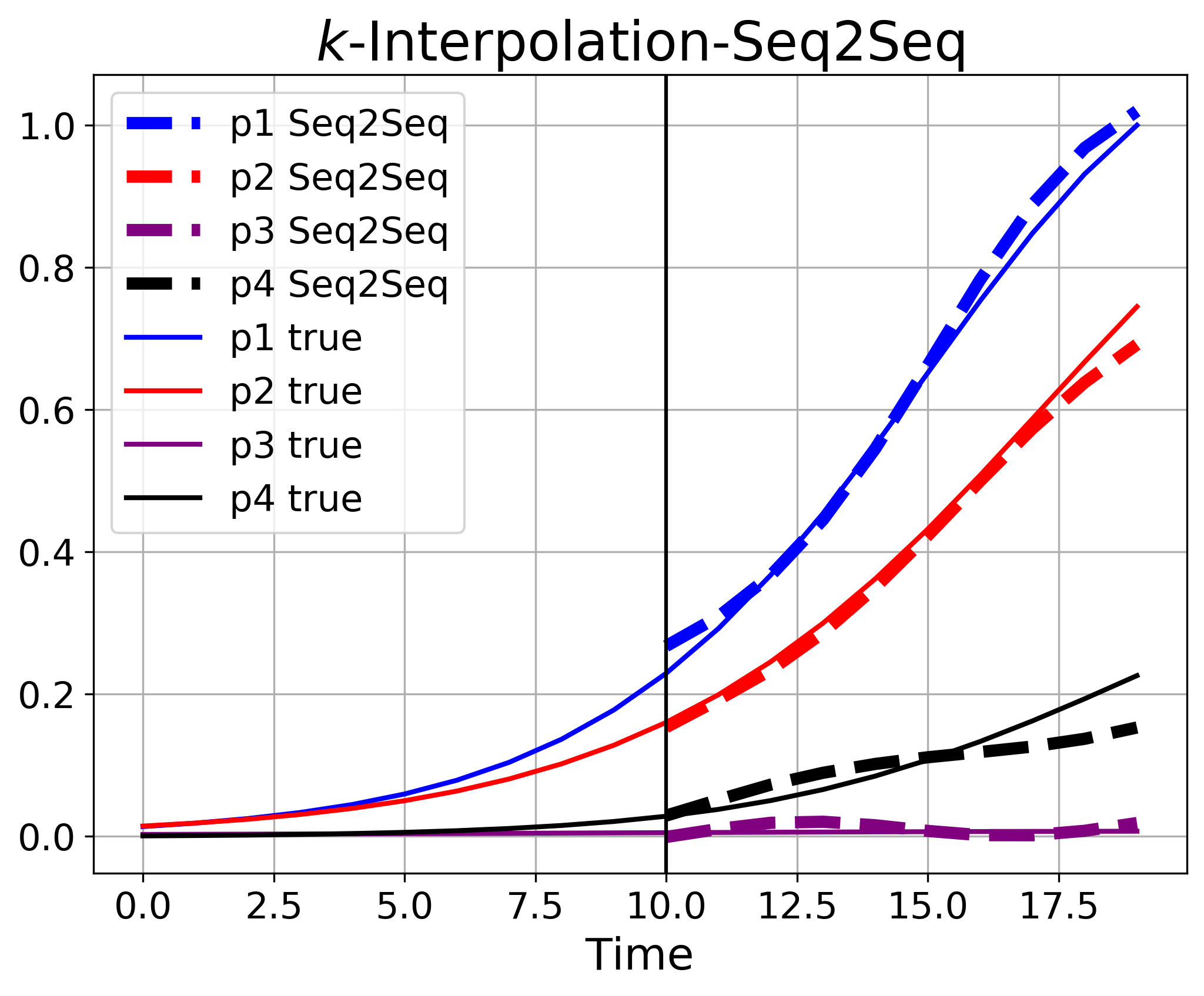}
\includegraphics[width=0.24\textwidth]{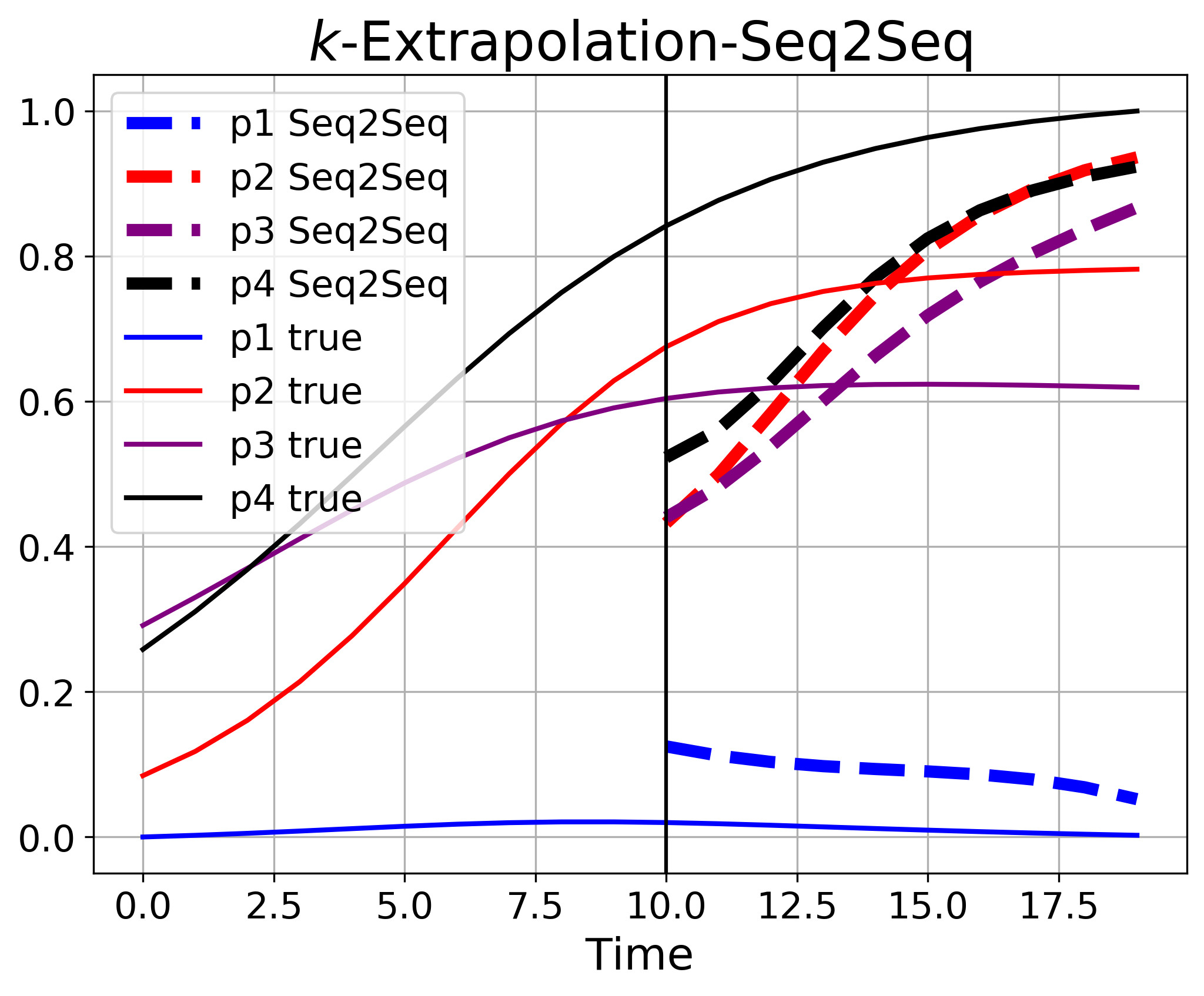}
}{
\vspace{-10pt}
\caption{\texttt{Seq2Seq} predictions on a $\bm{k}$-interpolation and a $\bm{k}$-extrapolation test samples of \textit{LV} dynamics, the vertical black line separates the input and forecasting period.}\label{fig_vis_lv}
\vspace{-.25cm}}
\ffigbox[0.49\textwidth]
{%
\includegraphics[width=0.24\textwidth]{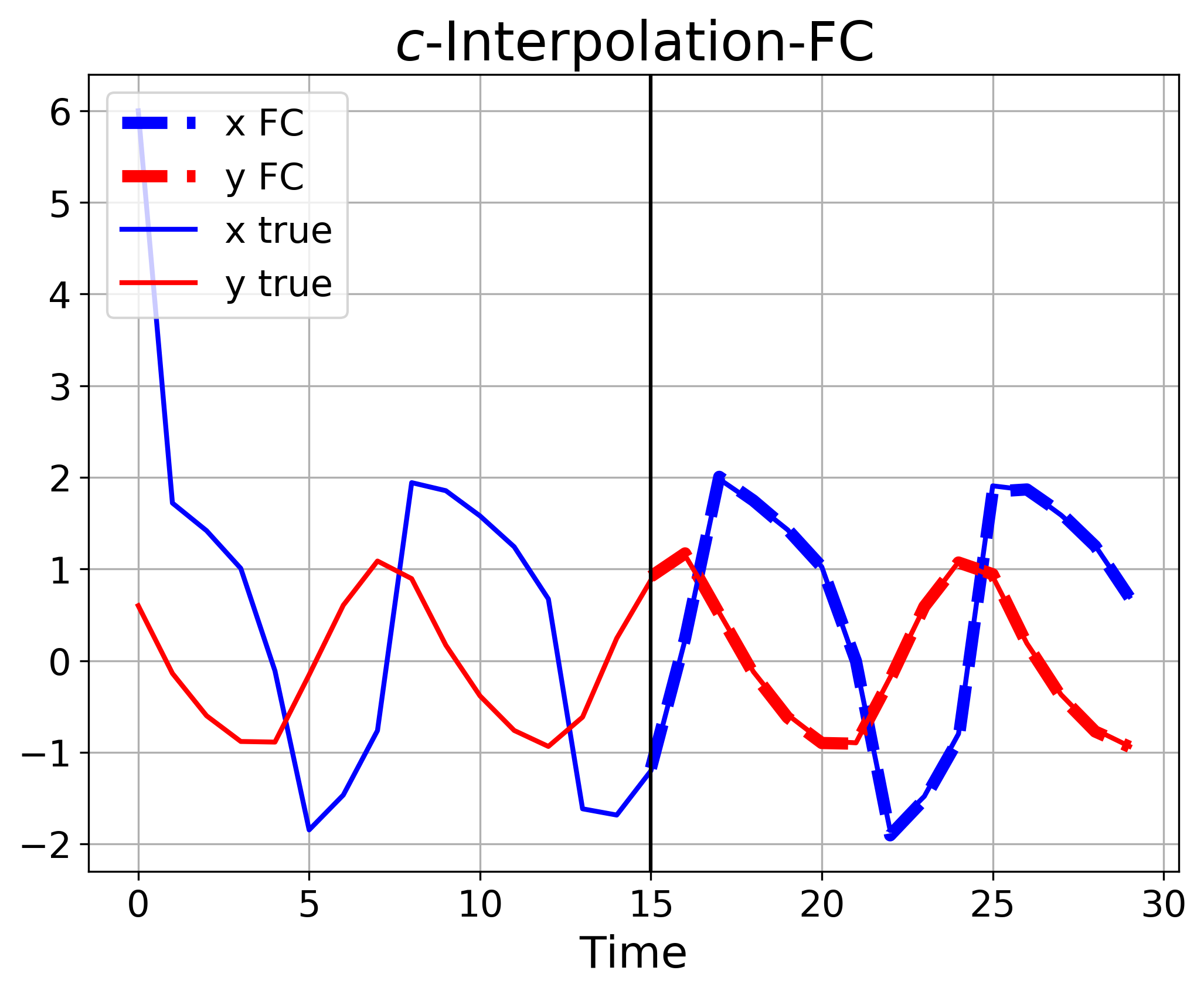}
\includegraphics[width=0.24\textwidth]{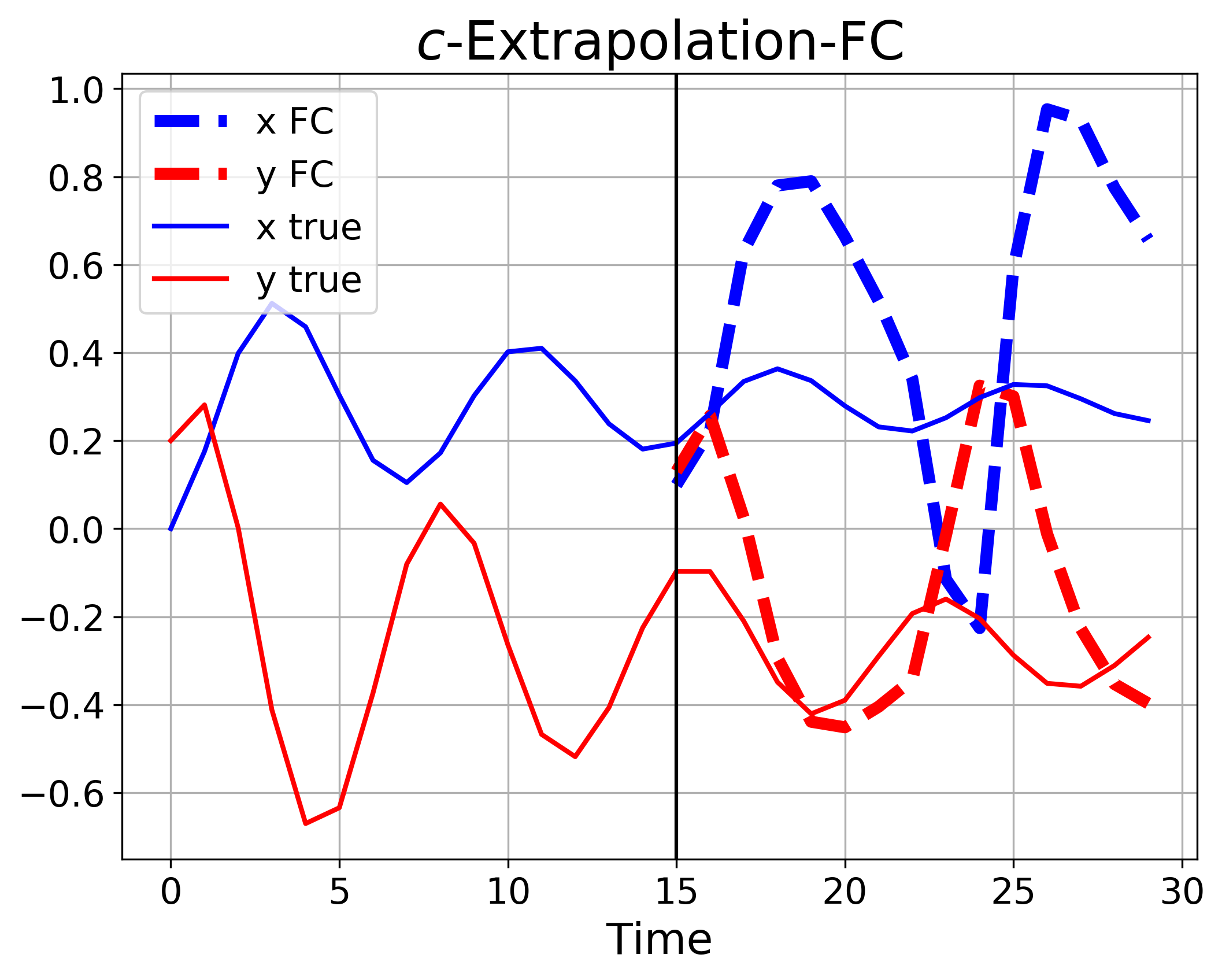}
}{
\vspace{-10pt}
\caption{\texttt{FC} predictions on a $c$-interpolation and a $c$-extrapolation test samples of \textit{FHN} dynamics, the vertical black line in the plots separates the input and forecasting period.}\label{fig_vis_fhn}
\vspace{-.3cm}}
\end{floatrow}
\vspace{-5mm}
\end{figure}

Table \ref{test_rmse} shows the prediction RMSEs of the models on initial values and system parameter interpolation and extrapolation test sets. We observe that the models' prediction errors on extrapolation test sets are much larger than the error on interpolation test sets. Figures \ref{fig_vis_lv}-\ref{fig_vis_fhn} show that \texttt{Seq2Seq} and \texttt{FC} fail to make accurate prediction when tested outside of the parameter distribution  even though they make accurate predictions for parameter interpolation test samples. 

\texttt{AutoODE} always obtains the lowest errors as it would be not affected by the range of parameters or the initial values. However, it is a local method and we need to train one model for each sample. 
In contrast, DL models can only mimic the behaviors of SEIR, LV and FHN dynamics rather than understanding the underlying mechanisms.


\vspace{-5pt}
\section{Conclusion}
We study the problem of forecasting non-linear dynamical systems.  From a benchmark study of DL  and physics-based models,  we find that \texttt{FC} and \texttt{Seq2Seq} have better prediction accuracy on the number of deaths, while the DL methods generally much worse  than the physics-based models on the number of infected and removed cases. This is mainly due to the distribution shift in the COVID-19 dynamics. On several other non-linear dynamical systems, we experimentally show that four DL models fail to generalize under shifted distributions in both the data and the parameter domains. Even though these models are powerful enough to memorize the training data, and perform well on the interpolation tasks. Our study provides important insights on learning real world dynamical systems: to achieve accurate forecasts with DL, we need to ensure that both the data and dynamical system parameters in the training set can cover the domains of the test set. Future works include incorporating compartmental models into deep learning models and derive theoretical generalization bounds of dynamics forecasting for deep learning models. 

\newpage
\bibstyle{apalike}
\bibliography{references.bib}

\end{document}